\documentclass[10pt,twocolumn,letterpaper]{article}

\usepackage{iccv}
\usepackage{times}
\usepackage{epsfig}
\usepackage{graphicx}
\usepackage{amsmath}
\usepackage{amssymb}
\usepackage{booktabs}
\usepackage{float}
\usepackage[ruled,linesnumbered]{algorithm2e}
\usepackage[accsupp]{axessibility}

\usepackage{multirow}

\usepackage[breaklinks=true,bookmarks=false]{hyperref}

\iccvfinalcopy 

\def\iccvPaperID{****} 

\ificcvfinal\fi
\begin{document}

\title{Hard-sample Guided Hybrid Contrast Learning \\ for Unsupervised Person Re-Identification}

\author{\textbf{Zheng Hu$^1$, Chuang Zhu$^{1}$\thanks{the corresponding author: Chuang Zhu (czhu@bupt.edu.cn)}, Gang He$^1$}\\
$^1$Beijing Laboratory of Advanced Information Networks\\
Beijing Key Laboratory of Network System Architecture and Convergence\\
Beijing University of Posts and Telecommunications\\

Beijing 100876, China\\
{ \{huzheng95, czhu, brianhe\}@bupt.edu.cn}
}

\maketitle
\ificcvfinal\thispagestyle{empty}\fi

\begin{abstract}
  Unsupervised person re-identification (Re-ID) is a promising and very challenging research problem in computer vision. Learning robust and discriminative features with unlabeled data is of central importance to Re-ID. Recently, more attention has been paid to unsupervised Re-ID algorithms based on clustered pseudo-label. However, the previous approaches did not fully exploit information of hard samples, simply using cluster centroid or all instances for contrastive learning. In this paper, we propose a Hard-sample Guided Hybrid Contrast Learning (HHCL) approach combining cluster-level loss with instance-level loss for unsupervised person Re-ID. Our approach applies cluster centroid contrastive loss to ensure that the network is updated in a more stable way. Meanwhile, introduction of a hard instance contrastive loss further mines the discriminative information. Extensive experiments on two popular large-scale Re-ID benchmarks demonstrate that our HHCL outperforms previous state-of-the-art methods and significantly improves the performance of unsupervised person Re-ID. The code of our work and dataset are available soon at \url{https://github.com/bupt-ai-cz/HHCL-ReID}.
\end{abstract}

\noindent{\bf Keywords:} {Unsupervised Learning; Person Re-ID; Hard Sample; Contrastive Learning, Pseudo Label. \\
}

\section{Introduction}

Person Re-ID aims to identify the same person under different cameras views. It has been used extensively in large-scale surveillance systems. Though great progress has been made in supervised person Re-ID tasks, the reliance on extensive manual annotation greatly constrains its application. Nevertheless, collecting pedestrian images without annotation is much cheaper and easier. Thus, increasing research attention has been drawn to unsupervised person Re-ID, directly learning from unlabeled data, which is more scalable and has more potential to deployments in the real world.

\begin{figure}[t]
  \centering
  \includegraphics[width=3.2in]{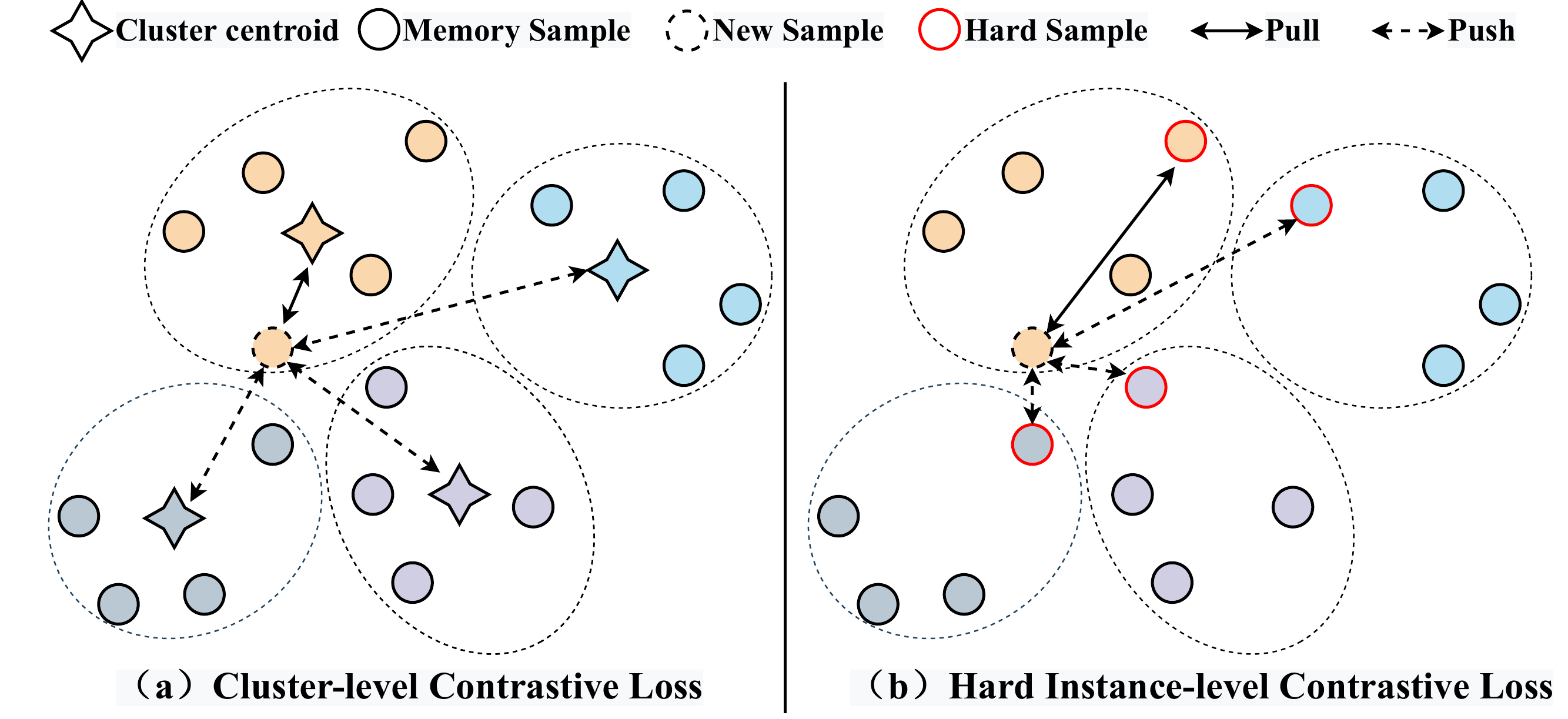}
\caption{{Hard-sample guided hybrid contrast learning. According to the features saved in the memory bank, we calculate cluster-level contrastive loss and hard instance-level contrastive loss, respectively. (a) Cluster centroid leads the optimization trend of features, resulting in features belonging to the same cluster being more compact and strengthen identity similarity. (b) Hard instance contrastive loss compares input sample with hard positive that belong to the same cluster and hard negative instances from other clusters, thereby learning to distinguish easily confusing samples. (Best viewed in color)}}
  \label{fig:hard instance mining}
\end{figure}

The extant unsupervised person re-ID methods can be broadly divided into two categories, unsupervised domain adaptation Re-ID methods and purely unsupervised Re-ID methods. The first type methods are based on unsupervised domain adaption (UDA) where the source domain dataset is fully annotated and the target domain is an unlabeled dataset. Most of these UDA-based methods address this task by learning the knowledge in the labeled source domain dataset and transferring them to the unlabeled target domain dataset \cite{MEB, DCML, MMT}.  The second type of unsupervised Re-ID method is pseudo-label-based fully unsupervised learning that directly learn from unlabeled data in the target domain and use representation features to estimate pseudo labels \cite{Wang2016TowardsUO, Yu2017CrossViewAM, SpCL}. This method does not require any annotations and is more challenging. Existing fully unsupervised Re-ID works mainly aim to exploit pseudo labels from clustering and apply contrastive learning which has shown excellent performance in unsupervised representation learning \cite{CL1, SimCLR, MoCo}. 

The performance of the unsupervised methods relies on feature representation learning. More recently, the State-of-the-art method \cite{MoCo} using a memory bank unit \cite{MemoryBank} to store all instance features, treats each image as an individual class, and learns the representation by matching features of the same instance in different augmented views. However, each class usually contains more than one positive instance in Re-ID datasets. SpCL \cite{SpCL} method alleviates this problem by matching an instance with the centroid of the multiple positives. To further ensure each positive converges to its centroid at a uniform pace, cluster contrast learning \cite{ClusterContrast} updates the memory dictionary and computes contrastive loss in the cluster level.

Although cluster contrast learning \cite{ClusterContrast} has achieved impressive performance, the method of applying contrastive learning only in the cluster level does not consider the the relationship between hard instances in the instance level. In fact, previous works in deep metric learning  have focused on hard sample mining to lay more emphasis on hard samples inside a class. These methods aim to distinguish samples from different categories and bring samples from the same category closer together. However, these methods usually adopt a mini-batch-based deep metric loss, such as hard triplet loss \cite{HardTripletLoss} and multi-similarity loss \cite{MSLoss}. Meanwhile, these losses only utilized a small portion of data without considering the information of all categories.

To learn discriminative feature representation for Re-ID and address the lack of adequately exploring information of hard samples, this paper introduces a novel hard-sample mining strategy and proposes a simple and effective method of hard-sample guided hybrid contrast learning for unsupervised Re-ID. In summary, this paper makes the following contributions:
 
\begin{itemize}
\item We propose a hybrid contrast learning framework for unsupervised person Re-ID which combines both cluster-level contrastive loss and instance-level contrastive loss.

\item We introduce a novel hard instance mining strategy, which is based on an instance memory bank, to explore more discriminative information by selecting global hard samples online for each input instance.

\item Extensive experiments on two popular large-scale Re-ID benchmarks demonstrate that our HHCL outperforms previous state-of-the-art methods and significantly improves the performance of unsupervised person Re-ID.
\end{itemize}


\section{Related Works} \label{section:2}
\subsection{Unsupervised Re-ID}


The domain adaptation strategy has been widely used for unsupervised person Re-ID tasks \cite{DCML, MMT}. The transfer-based methods follow the strategy of UDA, which uses the pre-trained model in the labeled source domain dataset as the initialization of the target domain, or uses the style transfer method to transfer labeled images to the target domain. However, the UDA approach can be very challenging when the categories in the two domains are quite different. The drawback with pseudo-labels is that if the domains are not similar enough, it is not easy for us to obtain high quality pseudo labels, because the labeling noise might be too high to hurt the performance. 


More recently, researchers have given more attention to pseudo-label-based methods that do not require source domain data. The pseudo labels can be generated by a pre-trained classifier or by a feature similarity-based clustering algorithm, such as K-means, DB-SCAN \cite{DB-SCAN}. In this way, the pseudo labels are applied to fine-tuning the Re-ID model in a supervised manner. HCT \cite{HCT} combined hierarchical clustering with hard-batch triplet loss to improve the quality of pseudo labels. MMCL \cite{MMCL} formulated unsupervised person re-ID as a multi-label classification task to progressively seek true labels. SpCL \cite{SpCL} adopted the self-paced contrastive learning strategy to form more reliable clusters. CACL \cite{AsymCL} designed an asymmetric contrastive learning framework to help the siamese network effectively mine the invariance in feature learning.


\subsection{Mining Schemes} 
Sampling is a fundamental operation for reducing bias during model learning. Random sampling is one of the commonly used approaches, and different sampling methods are proposed to facilitate the learning of various loss functions. For the person re-ID task, identity sampling is widely used during the training stage, such as pair-wise sampling for contrastive loss and semi-hard negative mining method for triplet loss. 

Hard sample mining is considered as a vital component of many deep metric learning algorithms \cite{MemoryBank} to accelerate network convergence or to improve the final discriminative ability of the neural network because hard samples are more informative for training. The training should focus more on hard samples than easy samples. However, existing hard mining schemes of deep metric learning based on mini-batch training data often suffer from slow convergence, because they employ only one negative or partial negative example in mini-batch while not interacting with the other negative classes that have not been sampled into the current mini-batch in each update. In this paper, we propose a new strategy selecting the global hard samples from a memory bank for each input feature, to improve the model performance. Our hard mining strategy considers the relationship between each query instance and other clusters of different pseudo labels rather than taking into account only the inter-instance relationship with a small fraction of the categories.

\begin{figure*}[htbp]
  \centering
  \includegraphics[width=6.46in]{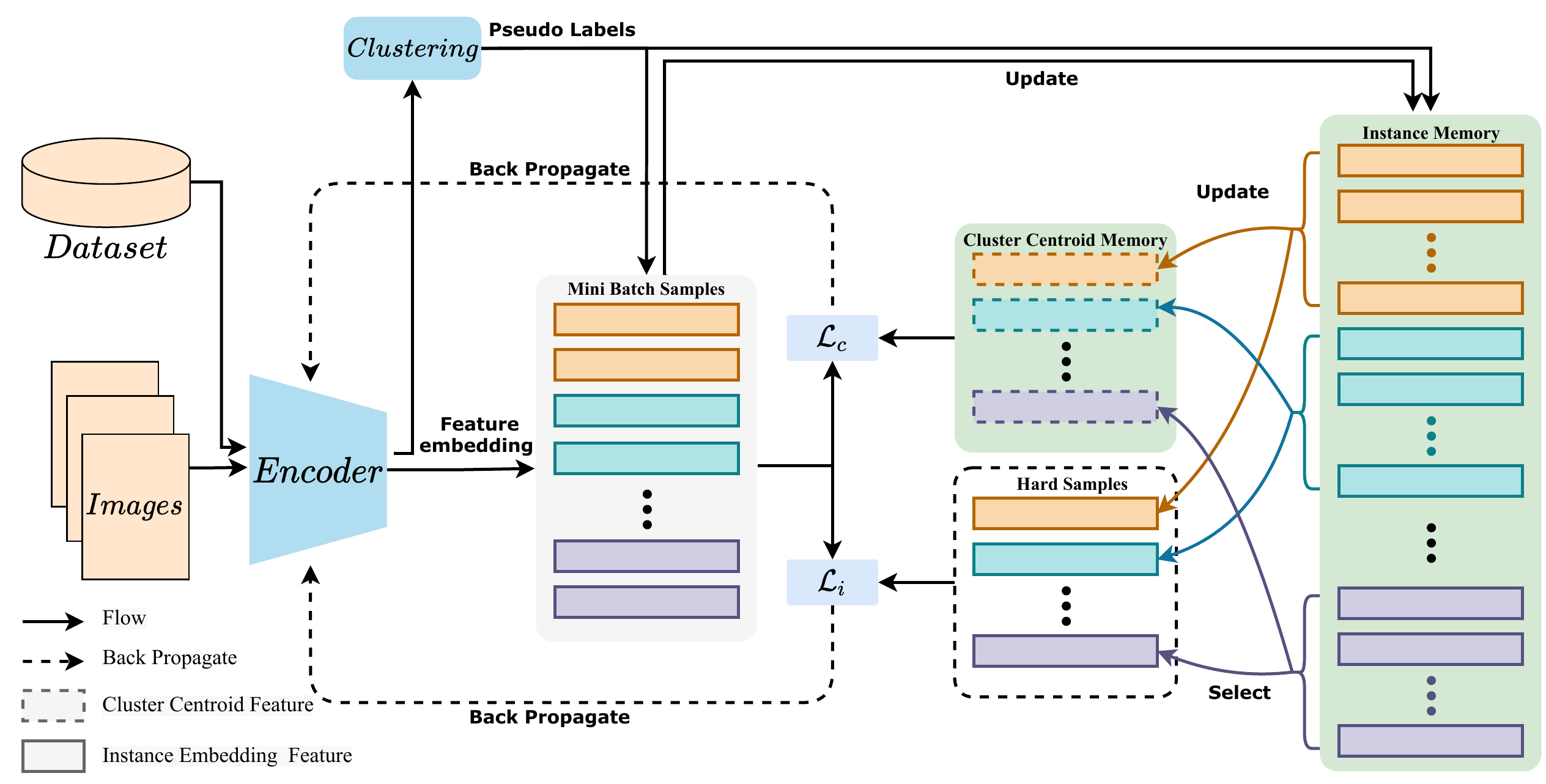}
  \caption{Hybrid Contrast Learning Framework. 1) Initialization: clustering algorithm divides all features extracted from the training set into different clusters as pseudo labels and initialize instance memory bank and cluster centroid memory bank; 2) forward propagate: calculate cluster contrastive loss between the input and the clustering centroids and the hard instance contrastive loss of the hard samples selected by hard mining strategy respectively; 3) back propagate and update the encoder model; 4) update instance memory bank and cluster centroid memory bank. }
  \label{fig:Framework}
\end{figure*}
\section{Preliminaries}
Given an unlabeled training set $\mathcal{X} = \{x_1, x_2,...,x_n\}$ consisting of $n$ image samples, the goal is to learn $\phi(\cdot; \theta)$—an encoder parameterized by $\theta$ used to extract features from input images. For inference, this encoder is applied to the gallery set $\mathcal{X}^g=\{x^g_1, x^g_2,...,x^g_{n_g}\}$ and query set $\mathcal{X}^q=\{x^q_1,x^q_2,...,x^q_{n_q}\}$. The gallery set contains the total collection of retrieval images in the database and representations of the query images $\phi(x^q_i; \theta)$ are used to search the gallery set to retrieve the most similar matches to $x^q_i$ according to Euclidean distance between the query and gallery embeddings, 
$d(x^q, x^g) = ||\phi(x^q; \theta) - \phi(x^g; \theta)||$, 
where a smaller distance implies increased similarity between the images. Thus, feature representations of the same person are supposed to be as close as possible. 

\section{Method}

\subsection{Architecture}

Our hybrid contrast learning framework for fully unsupervised Re-ID consists of two main components: Cluster Centroid Contrastive Loss (CCCL) and Hard Instance Contrastive Loss (HICL). As shown in Fig.\ref{fig:Framework}.

\subsection{Hybrid Contrast Learning} \label{Sec_Loss}
To increase intra-class compactness and inter-class separability, {state-of-the-art contrastive learning methods  minimize the distance between samples of the same category and maximize the distance between samples of different categories with InfoNCE loss \cite{InfoNCE}. }

\begin{equation*}
    \mathcal{L}_{q}=\mathbb{E}[
        -\log\frac{\exp(q \cdot k^{+})/\tau}{\sum_{i=1}^{K}\exp(q \cdot k^i)/\tau}], 
        \tag{1} 
    \label{eq:InfoNCE-Loss}
\end{equation*}
where $q$ is an encoded query and $k^+$ is a positive feature which has the same label with $q$ selected from a set of candidates ${k^1, k^2, ... k^K}$. $\tau$ is a temperature hyper-parameter that controls the scale of similarities.

Comparing the non-parametric loss functions of different approaches based on the memory dictionary, the SSL \cite{SSL} considers each image as an individual instance and computes the loss and updates the memory dictionary both in the instance level so that all features of the training data need to be saved. To decrease memory usage and take full advantage of clustering outliers, SPCL \cite{SpCL} computes the loss in cluster level but updates the memory dictionary in the instance level. However, the updating progress for each cluster is inconsistent due to the varying cluster size and randomness of sampling. ClusterNCE loss \cite{ClusterContrast} updates the feature vectors and computes the loss both in the cluster level. Although only a smaller storage space needs to be created to hold a cluster size amount of features for ClusterNCE, a single feature vector is not enough for a cluster representation. The averaged momentum representations calculated from all instances belonging to one cluster may lose the intra-class diversity. If updating cluster representation with only an instance feature, would introduce more biases because of noisy pseudo labels generated by unsupervised clustering.

Thus, we proposed a new unsupervised Re-ID framework that combines cluster-level loss with instance-level loss. The overall loss function of our method is as follow:
\begin{equation*}
    \mathcal{L}_{ReID}=\mu\mathcal{L}_{cls} + (1 - \mu)\mathcal{L}_{ins}, 
        \tag{2} 
    \label{eq:Total-Loss}
\end{equation*}
where $\mu$ is a balancing factor and we set $\mu$ = 0.5 by default. In the following, we will detail the objective function Eq.(\ref{eq:Total-Loss}).

\textbf{Cluster Centroid Contrastive Loss}
Some instance-level memory dictionary techniques, such as \cite{MMCL, MMT} maintaining each instance feature of the dataset and update corresponding memory dictionary with its own instance features in each mini-batch, have the problem of memory updating consistency \cite{ClusterContrast}. Since different instances within the same cluster will have different updating states. In every training iteration, due to the unbalanced distribution of cluster size, a smaller cluster could have a higher proportion of instances updated than a larger cluster. Unlike the previous instance-level memory dictionary, we use cluster-level memory dictionary $\mathcal{M}_{cls}$ to keep one cluster feature for each cluster instead of preserving every instance feature. The corresponding memory dictionary is updated regardless of whether the clusters are large or small, ensuring updating consistency of features within the same cluster.

\begin{equation*}
    \mathcal{L}_{cls}=\mathbb{E}[
        -\log\frac{\exp(<q \cdot c^{+}>)/\tau_c}{\sum_{i=1}^{C}\exp(<q \cdot c^i>)/\tau_c}], 
        \tag{3} 
    \label{eq:CC-Loss}
\end{equation*}
where $C$ is the number of clusters in a training epoch and $\tau_c$ is a temperature hyper-parameter. Different from unified contrastive loss, outliers are dropped out.


We calculate cluster centroids $c^1, c^2, ... c^{C} $and store them in a memory for the cluster centroid contrastive loss. We update the cluster memory bank as follows:
\begin{equation*}
    {c^i \leftarrow \alpha c^i + (1-\alpha) \bar{c}^i}, 
        \tag{4} 
    \label{eq:Update Cluster Memory}
\end{equation*}
where $\bar{c}^i$ is the average of $i$-th class instance features in the mini-batch.

\subsection{Memory Based Hard Mining Scheme}

To further distinguish easily confused sample pairs and explore the inter-instance relationship, we propose a novel hard sample mining strategy based on a memory dictionary. We construct another memory-based dictionary $\mathcal{M}_{ins}$ to store $P \times C$ instance features, which contains $C$ pseudo identities and each identity has $P$ instances. As shown in Fig.\ref{fig:hard instance mining}, unlike traditional hard mining strategies such as hard triplet loss \cite{HardTripletLoss}, which is based on pairwise loss calculating the distance of the hardest positive and the hardest negative instances within a mini-batch, our proposed method is based on all pseudo-labeled categories and contains $C-1$ negative samples for each query. Our hard mining strategy considers the relationship between each query instance and other clusters of different pseudo labels rather than taking into account only the inter-instance relationship with a small fraction of the categories. 

For the same query, we construct $C$ sample pairs which include one positive pair and $C-1$ hard negative pairs. We define hard instance contrastive loss as follows:
\begin{equation*}
    \mathcal{L}_{ins}=\mathbb{E}[
        -\log\frac{\exp(<q \cdot z^{+}_{hard}>)/\tau_{ins}}{\sum_{i=1}^{C}\exp(<q \cdot z^i_{hard}>)/\tau_{ins}}], 
        \tag{5} 
    \label{eq:HIC-Loss}
\end{equation*}
where $\tau_{ins}$ is an instance temperature hyper-parameter, $z^+_{hard}$ is the hard positive instance feature that has the lowest cosine similarity with query $q$ within the same cluster, and $z^i_{hard}$ is hard negative instance feature that has the highest cosine similarity that belongs to $i$-th class. They are respectively defined as
\begin{equation*}
    {z^+_{hard} = argmin(<q \cdot z^+_{k}>), {k = 1,...,K}};
        \tag{6} 
    \label{eq:Select Hard Pos Instance}
\end{equation*}
\begin{equation*}
    {z^i_{hard} = argmax(<q \cdot z^i_{k}>), {k = 1,...,K}}.
        \tag{7} 
    \label{eq:Select Hard Neg Instance}
\end{equation*}

Similarly, to ensure memory updating consistency, all instance features of the corresponding K identities in the mini-batch are updated in each training iteration. We update the instance memory bank as follows:
\begin{equation*}
    {m^i_{k} \leftarrow z^i_k}, 
        \tag{8} 
    \label{eq:Update Hard Memory}
\end{equation*}


\begin{algorithm}[h]\label{algo:1}
\caption{Hard-sample Guided Hybrid Contrast Learning for Unsupervised Re-Identification}
\label{alg:algorithm}
\KwData{An unlabeled training set $\mathcal{X}$}
\KwIn{ImageNet pre-trained model $\phi(\cdot; \theta)$, 
      the iteration number $N$,
      the training batch size $B$.}
\KwResult{trained model $\phi(\cdot; \theta)$}

\For{$epoch=1$ to $N$}{
    Extract feature embedding $x$ from $X$ by $\phi(\cdot; \theta)$\; 
    Clustering $x$ into $c$ clusters with DB-Scan\; 
    Initialize cluster memory bank $\mathcal{M}_{cls}$ and hard instance memory bank $\mathcal{M}_{ins}$ with Eq.1\;
\For{$iter=1$ to $B$}{
    Sample $P \times K$ mini-batch images from $\mathcal{X}$\;
    Forward to extract the features of the samples\;
    Compute the total loss $\mathcal{L}_{ReID}$ in Eq.(\ref{eq:Total-Loss}) which combines cluster centroid contrastive loss $\mathcal{L}_{cls}$ Eq.(\ref{eq:CC-Loss}) and hard instance contrastive loss $\mathcal{L}_{ins}$ Eq.(\ref{eq:HIC-Loss})\;
    Backward to update model $\phi_{\theta}$\;
    Update cluster memory $\mathcal{M}_{cls}$ and instance memory $\mathcal{M}_{ins}$ via Eq.(\ref{eq:Update Cluster Memory}) and Eq.(\ref{eq:Update Hard Memory})\;
    }
}
\end{algorithm}


\section{Experiments}
\subsection{Data and Metrics}
We evaluate our approach on two large-scale benchmark datasets: Market1501 \cite{Market1501}, and DukeMTMC-reID \cite{DukeMTMC} which are widely used real-world person Re-ID tasks.

\textbf{Market1501} contains 1,501 person identities with 32,668 images which are captured by 6 cameras in front of the Tsinghua University campus. It contains 12,936 images of 751 identities for training and 19,732 images of 750 identities for testing. All of the images were cropped by a pedestrian detector which inevitably introduced little misalignment, part missing and false positives.

\textbf{DukeMTMC-reID} consists a total of 36,411 images of people from 1404 different identities collected by 8 cameras. Specifically, The dataset is split by randomly selecting 702 identities as the training set and 702 identities as the testing set. it contains 16,522 images for training, 2,228 query images and 17,661 gallery images for testing.


\textbf{Evaluation Metrics.} We followed the standard training/test split and evaluation protocol to evaluate the performance of our method. For the evaluation metrics, we used the Rank-k (for k = 1, 5, and 10) matching accuracy, which means the query picture has the match in the top-k list. And we use the mean Average Precision (mAP), which is computed from the Cumulated Matching Characteristics (CMC) \cite{CMCEvaluatingAM}. Moreover, results reported in this paper are under the single-query setting, and no post-processing technique is applied.

\subsection{Implementation}
We adopt ResNet-50 \cite{Resnet} as the backbone of the feature extractor and initialize the model with the parameters pre-trained on ImageNet \cite{ImageNet}. After layer-4, we remove all sub-module layers and add global average pooling (GAP) followed by batch normalization layer \cite{BN} and L2-normalization layer, which will produce 2048-dimensional features. During testing, we take the features of the global average pooling layer to calculate the distance. For the beginning of each epoch, we use DB-SCAN \cite{DB-SCAN} for clustering to generate pseudo labels. The input image is resized $256 \times 128$. For training images, we perform random horizontal flipping, padding with 10 pixels, random cropping, and random erasing. Each mini-batch contains 256 images of 16 pseudo person identities (16 instances for each person). We adopt Adam optimizer to train the Re-ID model with weight decay 5e-4. The initial learning rate is set to 3.5e-4, and is reduced to 1/10 of its previous value every 20 epoch in a total of 50 epoch. As the same with the cluster method of \cite{SpCL} paper, we use DB-SCAN and Jaccard distance \cite{RerankingPR} to cluster with k nearest neighbors, where k = 30. For DB-SCAN, the maximum distance d between two samples is set as 0.45 and the minimal number of neighbors in a core point is set as 4.

\subsection{Results}
\subsubsection{Comparison with unsupervised method}

\begin{table*}[htb]
\centering 
\begin{tabular}{llllllllll}
\hline
\multicolumn{1}{l|}{\multirow{2}{*}{Method}} & \multicolumn{1}{l|}{\multirow{2}{*}{Reference}} & \multicolumn{4}{c|}{Market1501} & \multicolumn{4}{c}{DukeMTMC-reID} \\ \cline{3-10} 
\multicolumn{1}{l|}{} & \multicolumn{1}{l|}{} & \multicolumn{1}{c}{mAP} & \multicolumn{1}{c}{R1} & \multicolumn{1}{c}{R5} & \multicolumn{1}{c|}{R10} & \multicolumn{1}{c}{mAP} & \multicolumn{1}{c}{R1} & \multicolumn{1}{c}{R5} & \multicolumn{1}{c}{R10} \\ \hline
\multicolumn{10}{l}{Unsupervised Domain Adaptation} \\ \hline
\multicolumn{1}{l|}{ECN \cite{ECN}} & \multicolumn{1}{l|}{CVPR'19} & 43.0 & 75.1 & 87.6 & \multicolumn{1}{l|}{91.6} & 40.4 & 63.3 & 75.8 & 80.4 \\
\multicolumn{1}{l|}{MAR\cite{MAR}} & \multicolumn{1}{l|}{CVPR'19} & 40.0 & 67.7 & 81.9 & \multicolumn{1}{c|}{-} & 48.0 & 67.1 & 79.8 & \multicolumn{1}{c}{-} \\
\multicolumn{1}{l|}{SSG\cite{SSG}} & \multicolumn{1}{l|}{ICCV'19} & 58.3 & 80.0 & 90.0 & \multicolumn{1}{l|}{92.4} & 53.4 & 73.0 & 80.6 & 83.2 \\
\multicolumn{1}{l|}{MMCL \cite{MMCL}} & \multicolumn{1}{l|}{CVPR'20} & 60.4 & 84.4 & 92.8 & \multicolumn{1}{l|}{95.0} & 51.4 & 72.4 & 82.9 & 85.0 \\
\multicolumn{1}{l|}{JVTC \cite{JVTC}} & \multicolumn{1}{l|}{ECCV'20} & 61.1 & 83.8 & 93.0 & \multicolumn{1}{l|}{95.2} & 56.2 & 75.0 & 85.1 & 88.2 \\
\multicolumn{1}{l|}{DG-Net++ \cite{DG-Net++}} & \multicolumn{1}{l|}{ECCV'20} & 61.7 & 82.1 & 90.2 & \multicolumn{1}{l|}{92.7} & 63.8 & 78.9 & 87.8 & 90.4 \\
\multicolumn{1}{l|}{ECN+ \cite{ECN+}} & \multicolumn{1}{l|}{PAMI'20} & 63.8 & 84.1 & 92.8 & \multicolumn{1}{l|}{95.4} & 54.4 & 74.0 & 83.7 & 87.4 \\
\multicolumn{1}{l|}{MMT \cite{MMT}} & \multicolumn{1}{l|}{ICLR'20} & 71.2 & 87.7 & 94.9 & \multicolumn{1}{l|}{96.9} & 65.1 & 78.0 & 88.8 & 92.5 \\
\multicolumn{1}{l|}{DCML \cite{DCML}} & \multicolumn{1}{l|}{ECCV'20} & 72.6 & 87.9 & 95.0 & \multicolumn{1}{l|}{96.7} & 63.3 & 79.1 & 87.2 & 89.4 \\
\multicolumn{1}{l|}{MEB \cite{MEB}} & \multicolumn{1}{l|}{ECCV'20} & 76.0 & 89.9 & 96.0 & \multicolumn{1}{l|}{97.5} & 66.1 & 79.6 & 88.3 & 92.2 \\
\multicolumn{1}{l|}{SpCL \cite{SpCL}} & \multicolumn{1}{l|}{NeurIPS'20} & 76.7 & 90.3 & 96.2 & \multicolumn{1}{l|}{97.7} & 68.8 & 82.9 & 90.1 & 92.5 \\ \hline
\multicolumn{10}{l}{Fully Unsupervised} \\ \hline
\multicolumn{1}{l|}{SSL \cite{SSL}} & \multicolumn{1}{l|}{CVPR'20} & 37.8 & 71.7 & 83.8 & \multicolumn{1}{l|}{87.4} & 28.6 & 52.5 & 63.5 & 68.9 \\
\multicolumn{1}{l|}{JVTC \cite{JVTC}} & \multicolumn{1}{l|}{ECCV'20} & 41.8 & 72.9 & 84.2 & \multicolumn{1}{l|}{88.7} & 42.2 & 67.6 & 78.0 & 81.6 \\
\multicolumn{1}{l|}{MMCL \cite{MMCL}} & \multicolumn{1}{l|}{CVPR'20} & 45.5 & 80.3 & 89.4 & \multicolumn{1}{l|}{92.3} & 40.2 & 65.2 & 75.9 & 80.0 \\
\multicolumn{1}{l|}{HCT \cite{HCT}} & \multicolumn{1}{l|}{CVPR'20} & 56.4 & 80.0 & 91.6 & \multicolumn{1}{l|}{95.2} & 50.7 & 69.6 & 83.4 & 87.4 \\
\multicolumn{1}{l|}{CycAs \cite{wang2020cycas}} & \multicolumn{1}{l|}{ECCV'20} & 64.8 & 84.8 & \multicolumn{1}{c}{-} & \multicolumn{1}{c|}{-} & 60.1 & 77.9 & \multicolumn{1}{c}{-} & \multicolumn{1}{c}{-} \\
\multicolumn{1}{l|}{SpCL \cite{SpCL}} & \multicolumn{1}{l|}{NeurIPS'20} & 73.1 & 88.1 & 95.1 & \multicolumn{1}{l|}{97.0} & 65.3 & 81.2 & 90.3 & 92.2 \\
\multicolumn{1}{l|}{CAP \cite{CAP}} & \multicolumn{1}{l|}{AAAI'21} & 79.2 & 91.4 & 96.3 & \multicolumn{1}{l|}{97.7} & 67.3 & 81.1 & 89.3 & 91.8 \\
\multicolumn{1}{l|}{CACL \cite{AsymCL}} & \multicolumn{1}{l|}{Arxiv'21} & 80.9 & 92.7 & 97.4 & \multicolumn{1}{l|}{98.5} & 69.6 & 82.6 & 91.2 & 93.8 \\
\multicolumn{1}{l|}{ICE\cite{ICE}} & \multicolumn{1}{l|}{ICCV' 21} & 82.3 & \textbf{93.8} & 97.6 & \multicolumn{1}{l|}{98.4} & 69.9 & 83.3 & 91.5 & 94.1 \\
\multicolumn{1}{l|}{CCL\cite{ClusterContrast}} & \multicolumn{1}{l|}{Arxiv'21} & 82.6 & 93.0 & 97.0 & \multicolumn{1}{l|}{98.1} & 72.8 & \textbf{85.7} & 92.0 & 93.5 \\
\multicolumn{1}{l|}{HHCL} & \multicolumn{1}{l|}{This paper} & \textbf{84.2} & 93.4 & \textbf{97.7} & \multicolumn{1}{l|}{\textbf{98.5}} & \textbf{73.3} & \textbf{85.1} & \textbf{92.4} & \textbf{94.6} \\ \hline
\multicolumn{10}{l}{Supervised} \\ \hline
\multicolumn{1}{l|}{PCB\cite{PCB}} & \multicolumn{1}{l|}{ECCV'18} & 81.6 & 93.8 & \multicolumn{1}{c}{97.5} & \multicolumn{1}{c|}{98.5} & \multicolumn{1}{c}{69.2} & \multicolumn{1}{c}{83.3} & \multicolumn{1}{c}{90.5} & \multicolumn{1}{c}{92.5} \\
\multicolumn{1}{l|}{OSNet \cite{OSnet}} & \multicolumn{1}{l|}{ICCV' 19} & 84.9 & 94.8 & \multicolumn{1}{c}{-} & \multicolumn{1}{c}{-} & 73.5 & 88.6 & \multicolumn{1}{c}{-} & \multicolumn{1}{c}{-} \\
\multicolumn{1}{l|}{DG-Net \cite{DG-Net}} & \multicolumn{1}{l|}{CVPR'19} & 86.0 & 94.8 & \multicolumn{1}{c}{-} & \multicolumn{1}{c|}{-} & \multicolumn{1}{c}{74.8} & \multicolumn{1}{c}{86.6} & \multicolumn{1}{c}{-} & \multicolumn{1}{c}{-} \\
\multicolumn{1}{l|}{ICE \cite{ICE} (w/ GT)} & \multicolumn{1}{l|}{ICCV' 21} & 86.6 & \textbf{95.1} & 98.3 & \multicolumn{1}{l|}{98.9} & 76.5 & 88.2 & 94.1 & 95.7 \\
\multicolumn{1}{l|}{HHCL(w/ GT)} & \multicolumn{1}{l|}{This paper} & \textbf{87.2} & 94.6 & \textbf{98.5} & \multicolumn{1}{l|}{\textbf{99.1}} & \textbf{80.0} & \textbf{89.8} & \textbf{95.2} & \textbf{96.7} \\ \hline
\end{tabular}
\caption{Experimental results of the proposed HHCL and state-of-the-art methods on Market-1501 and DukeMTMC-reID. Note that the best results are bolded.}
\label{vs_SOTA}
\end{table*}

We compare our proposed method with state-of-the-art ReID methods including: 1) the unsupervised domain adaptation methods for person Re-ID(e.g. ECN \cite{ECN}, MAR\cite{ECN}, SSG\cite{SSG}, MMCL\cite{MMCL}, JVTC\cite{JVTC}, DG-Net++\cite{DG-Net++}, ECN+\cite{ECN+}, MMT\cite{MMT}, DCML\cite{DCML}, MEB\cite{MEB}, SpCL \cite{SpCL}; 2) the purely unsupervised methods for person Re-ID  SSL\cite{SpCL}, MMCL\cite{MMCL}, JVTC\cite{JVTC}, HCT\cite{HCT}, CycAs\cite{wang2020cycas}, SpCL\cite{SpCL}, CAP\cite{CAP}, CACL \cite{AsymCL}, CCL \cite{ClusterContrast} and ICE\cite{ICE}). The comparison results of the state-of-the-art unsupervised domain adaptation methods and purely unsupervised methods on Market-1501 and DukeMTMC-reID are reported in Tab. \ref{vs_SOTA}. 

As shown in Tab.\ref{vs_SOTA}, we observe our method is competitive with all the state-of-the-art methods. On the three datasets, our proposed HHCL without any identity annotation achieves better performance than all of UDA methods that use of the additional labeled source dataset.
It can be found that we not only perform better than all unsupervised domain adaptation methods and also achieve competitive performance with purely unsupervised methods. Under the fully unsupervised setting, HHCL achieves $84.2\%$ in mAP and $93.4\%$ in rank-1 accuracy on Market-1501, which is 1.9\% higher than the current state of the art (ICE \cite{ClusterContrast}). On DukeMTMC-reID, our method also achieves a high performance of $73.3/85.1\%$ in mAP/rank-1. These results indicate that our method is effective for unsupervised person Re-ID learning.

\subsubsection{Comparison with supervised method}

Our HHCL method can be easily implemented as a supervised approach when we replace the pseudo-labels with ground truth. We further find that our proposed unsupervised method is already comparable to some excellent supervised methods, such as PCB \cite{PCB} and DG-Net \cite{DG-Net}, when ground truth is not used. And our HHCL even achieves a better performance under supervised setting. This result shows that our proposed method achieves better results when using the ground truth to avoid introducing noisy pseudo-labels. And it also further demonstrates the effectiveness of our method for the person Re-ID problem, both unsupervised and supervised.

\subsection{Ablation Study}

\begin{table}[bp]
\centering
\begin{tabular}{c|cccc}
\hline
\multirow{2}{*}{$\mu$} & \multicolumn{4}{c}{Market1501} \\ \cline{2-5} 
& mAP & R1 & R5 & R10 \\ \hline
0(hard) & 78.5 & 90.5 & 96.0 & 97.4 \\
0.25 & 82.7 & 92.9 & 97.0 & 98.2 \\
0.5 & \textbf{84.2} & \textbf{93.4} & \textbf{97.7} & \textbf{98.5} \\
0.75 & 81.7 & 92.1 & 97.1 & 98.2 \\
1.(mean) & 80.8 & 91.3 & 96.3 & 97.6 \\ \hline
\end{tabular}
\caption{Evaluation of parameter $\mu$ on Market1501.}
\label{ablation}
\end{table}

\begin{figure}[htbp]
  \centering
\begin{minipage}[t]{0.4\textwidth}
  \centering
  \includegraphics[width=1.0\textwidth]{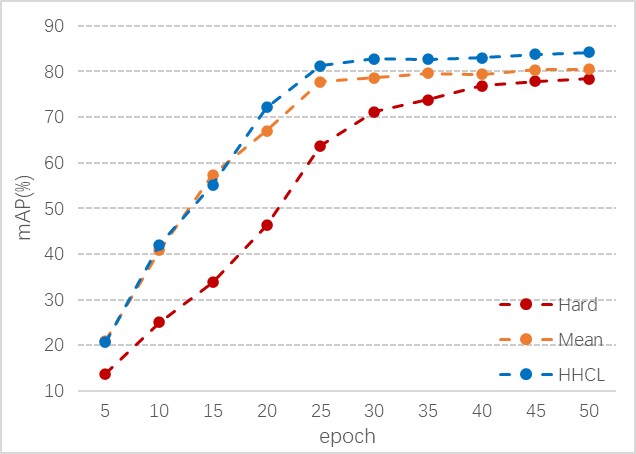}
  \label{fig:Ablation_mAP}
  \end{minipage}
  \hspace{.15in}
\begin{minipage}[t]{0.4\textwidth}
   \centering
  \includegraphics[width=1.0\textwidth]{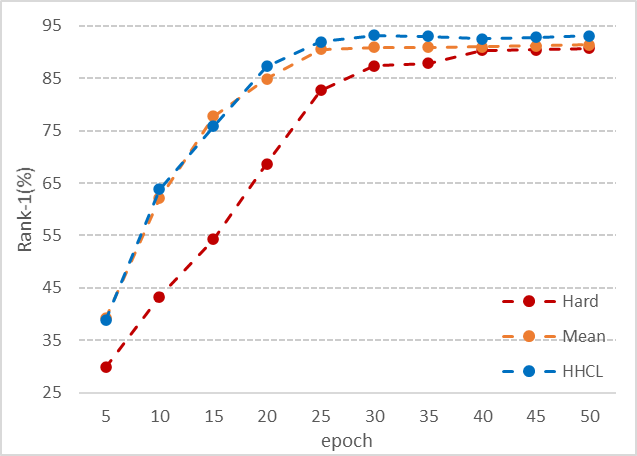}
  \caption{Ablation study on Market1501: Result comparisons of different settings in mAP and Rank-1.}
  \label{fig:Ablation_Rank1}
  \end{minipage}
\end{figure}

\textbf{Influence of Hyper-Parameter $\mu$} Tab. \ref{ablation} reports the experiment result under different value of hyper-parameter$\mu$. As mentioned in \ref{eq:Total-Loss}, $\mu$ is a balancing factor between 0 and 1, which plays an important role in affecting the weights of the cluster-level loss and instance-level loss. When $\mu$ is equal to 0, the loss function contains only the hard instance contrastive loss. From the fig.\ref{fig:Ablation_Rank1}. we can find that the model converges very slowly in the early stage of the training process, and using only the hard samples for comparison is not benefit for learning generalized features and obtaining better clustering pseudo labels. On the contrary, when $\mu$=1 and cluster-level loss only is used, although a faster convergence  can be achieved, only one feature is retained for each cluster, which loses the diversity of intra class and is still not conducive to facilitating the network to learn more discriminative features. It can be seen that combining both kind of contrastive loss leads to better performance obviously. And when $\mu$ = 0.5, we get the best performance 84.2\% in mAP, indicating that our proposed hybrid contrastive learning method has a distinct advantage over others during the training process.

\begin{table}[h]
\centering
\begin{tabular}{l|cccc}
\hline
\multirow{2}{*}{Method} & \multicolumn{4}{c}{Market1501} \\ \cline{2-5} 
 & mAP & R1 & R5 & R10 \\ \hline
ResNet50 & 84.2 & 93.4 & 97.7 & 98.5 \\
IBN + GeM & 87.8 & 95.1 & 98.2 & 98.8 \\
IBN + GeM + LS & 88.2 & 94.9 & 98.3 & 98.9 \\ \hline
\multirow{2}{*}{Method} & \multicolumn{4}{c}{DukeMTMC-reID} \\ \cline{2-5} 
 & mAP & R1 & R5 & R10 \\ \hline
ResNet50 & 73.3 & 85.1 & 92.4 & 94.6 \\
IBN + GeM & 76.8 & 87.9 & 93.4 & 94.9 \\
IBN + GeM + LS & 77.3 & 87.7 & 93.5 & 95.1 \\ \hline
\end{tabular}
\caption{Comparison of HHCL with other tricks on Market1501 and DukeMTMC-reID datasets. 'IBN' denotes that the backbone applies IBN-ResNet50. 'GeM' and 'LS' represent GeM pooling layer and label smoothing respectively. }
\label{tricks}
\end{table}
Instance-batch normalization (IBN) \cite{IBN} and  Generalized Mean Pooling (GeM) \cite{GeM} has been proved effective in both supervised and UDA based Re-ID methods. We compare the performance of HCCL under different settings in Tab.\ref{tricks}. The performance of our proposed HHCL can be further improved with an IBN-ResNet50 backbone network and GeM pooling layer.

\section{Conclusion}
In this paper, we propose a novel method for the fully unsupervised person re-ID. The new concepts and techniques introduced include a more efficient hybrid contrast learning framework and a memory based hard sample mining scheme. Specifically, our proposed HHCL approach comprehensively consider both of cluster level and instance level information. For effectively exploiting the invariance within and between clusters, HHCL leverages hard samples to guide network to learn more robust and discriminative features. Extensive experiments on two benchmark datasets demonstrated that HHCL achieves the best results comparing with all existing purely unsupervised and UDA-based Re-ID methods.

\section*{Acknowledgements}
This work was supported in part by 111 Project of China (B17007), and in part by the National Natural Science Foundation of China (61602011).

{\small
\bibliographystyle{ieee_fullname}
\bibliography{egbib}
}

\end{document}